\documentclass[12pt]{article}
\usepackage[a4paper,top=3.5cm,bottom=3.5cm,left=3.5cm,right=3.5cm]{geometry}

\usepackage{amssymb}
\usepackage{amsmath}

\usepackage{algorithm}
\usepackage{algpseudocode}
\usepackage{standalone} 
\usepackage{graphicx}   
\usepackage[T1]{fontenc}  
\usepackage[utf8]{inputenc}  
\usepackage{lmodern}  
\usepackage{longtable}
\usepackage{booktabs}
\usepackage{multirow}
\usepackage{siunitx}
\usepackage{natbib}
\usepackage{authblk}
\usepackage{tikz}
\usetikzlibrary{shapes.geometric, arrows, calc}

\usepackage{pgfplots}
\pgfplotsset{compat=1.18}

\newtheorem{definition}{Definition}

\title{Partial Column Generation with Graph Neural Networks for Team Formation and Routing}

\author[1]{Giacomo Dall'Olio}
\author[1]{Rainer Kolisch}
\author[2]{Yaoxin Wu}
\affil[1]{School of Management, Department of Operations \& Technology
Technical University of Munich, Germany, giacomo.dallolio@tum.de, rainer.kolisch@tum.de}
\affil[2]{Department of Information Systems, Eindhoven University of Technology, The Netherlands, ywu2@tue.nl}

\date{}

\begin{document}

\maketitle


\begin{abstract}
\noindent The team formation and routing problem is a challenging optimization problem with several real-world applications in fields such as airport, healthcare, and maintenance operations. 
To solve this problem, exact solution methods based on column generation have been proposed in the literature.
In this paper, we propose a novel partial column generation strategy for settings with multiple pricing problems, based on predicting which ones are likely to yield columns with a negative reduced cost.
We develop a machine learning model tailored to the team formation and routing problem that leverages graph neural networks for these predictions.
Computational experiments demonstrate that applying our strategy enhances the solution method and outperforms traditional partial column generation approaches from the literature, particularly on hard instances solved under a tight time limit. 
\end{abstract}

\vspace{3ex}

\section{Introduction}
\label{sec:Intro}
The team formation and routing problem consists of assigning skilled workers or technicians to teams and planning their routes to undertake tasks.
This problem is becoming increasingly popular and has various real-world applications such as non-emergency patient transport \citep{huang2024exact}, on-site maintenance \citep{cabrera2025workforce}, and home care \citep{bazirha2023efficient}.
In this paper, we focus on an airport operations application, i.e., the formation and routing of teams for baggage loading and unloading.
Given that the airport operations environment is highly dynamic and fluid, plans need frequent adjustments to adapt to changes.
For this reason, it is crucial for the planning algorithms to be fast.
\citet{hagn2024branch} have proposed an optimization algorithm based on branch-and-price to solve the problem.
{However, their approach does not exploit the fact that this problem must be solved multiple times each day and, therefore, it misses the opportunity to learn from previous runs to accelerate future executions. 
In our work, we address this by integrating Machine Learning (ML) into the Column Generation (CG) routine of the proposed solution method to reduce the overall computational time.
Specifically, we propose a novel partial column generation approach using Graph Neural Networks (GNN) to guide the selection of the pricing problems to solve.

Column generation (CG) is a popular technique to solve large linear programs \citep{desaulniers2006column}.
CG can efficiently solve linear programs that require only a small subset of variables to identify an optimal solution, e.g., formulations using the Dantzig-Wolfe decomposition.
To apply CG, the original problem is first formulated as a master problem (MP), which typically contains too many variables to handle directly.
CG works with the Restricted Master Problem (RMP), which is a linear program with the same constraints as the MP but defined on a much smaller subset of variables.
A feasible solution to the RMP is, therefore, feasible to the MP as well.
The reduced size of the RMP makes it generally easy to solve.
Given an optimal solution to the RMP, there might be other variables of the MP that can improve this solution if included in the RMP.
We know from the simplex algorithm that a variable has potential to improve the current solution only if its reduced cost is negative, assuming a minimization problem.
The pricing problem (PP) is the problem of finding a column in the MP that minimizes the reduced cost of the corresponding variable given the optimal dual solution to the RMP.
If there is a variable/column with a negative reduced cost, we can include it in the RMP and possibly find a better solution.
We iteratively repeat the procedure until we demonstrate that there are no more variables with a negative reduced cost.
This proves that the current optimal solution to the RMP is optimal to the MP as well.
To derive integer solutions, we can embed CG in a branch-and-price framework.

CG has successfully been used to solve several types of combinatorial problems \citep{lubbecke2005selected}, e.g., cutting stock problems, vehicle routing problems (VRPs), and crew scheduling problems (CSPs) .
In many practical applications, the PP can be divided into multiple PPs.
This makes sense when the columns are not all subject to the same constraints.
For a VRP with a heterogeneous fleet, e.g., we can define a PP for each type of vehicle.
In this way, each of the PPs can independently express the different constraints corresponding to the different types of vehicle.
The actual number of PPs strongly depends on the specific problem.

Finding the column with the minimal reduced cost is not strictly necessary at every iteration of the CG procedure.
As soon as we find any columns with a negative reduced cost, we can potentially add them to the RMP and solve it again.
One could achieve this by solving the PPs heuristically or by solving a fraction of them.
Since the literature on column generation is not always consistent on the terminology of these strategies, we state the one that we use in the remainder of this paper.
We call the strategy that solves all pricing problems to optimality \textit{Full Pricing} (FC) while we denote by \textit{Partial Pricing} any strategy that does not always solve all PPs or does not always solve them to optimality.
We refer to partial pricing strategies that focus on solving the PPs heuristically as \textit{Heuristic Pricing}.
Finally, we refer to partial pricing strategies that do not solve all PPs in every iteration of the CG as \textit{Partial Column Generation} (PCG).

Partial pricing is a known acceleration technique in the literature on CG.
The goal of partial pricing is to reduce the computational time of the individual CG iterations.
Note that a CG iteration is defined as a single cycle of solving the RMP and the PPs in the CG loop.
The potential drawback of partial pricing is that it could slow down the convergence of the CG loop, that is, we might need more iterations to terminate.
This can happen because we might miss the columns with the most negative reduced costs, which are those with the highest potential to improve the value of the current solution.
For partial pricing to induce an actual acceleration, a trade-off must be found between the speed-up of the CG iterations and their increased number.
Note that one must always resort to full pricing once partial pricing fails to produce any columns with a negative reduced cost, independently of the used strategy.
This is necessary for the algorithm to prove optimality.

In this paper, we aim to accelerate the solution method proposed to solve the team formation and routing problem with a new partial column generation approach, in which we dynamically select which PPs to solve at every CG iteration.
Unlike in other PCG strategies, we focus on identifying which PPs are useful instead of sorting them or selecting them in a round-robin fashion.
Specifically, our goal is to predict which of the PPs are likely to return a column with a negative reduced cost before actually solving them.
The idea is to solve only those PPs for which we are confident that they will contribute to the convergence of the CG.
We expect this to reduce the computation time of the CG iterations without significantly increasing their number.
The PPs in the proposed solution method constitute an Elementary Shortest Path Problem with Resource Constraints (ESPPRC), one of the most common types of PP for problems solved with CG.
To make predictions for our PCG strategy, we develop an ML model based on GNNs tailored for the specific variation of ESPPRC.
The choice of using GNNs is motivated by two key aspects.
First, GNNs are designed to operate directly on graph-structured data, which aligns naturally with the structure of ESPPRCs.
Second, GNNs can handle inputs of varying sizes, which is essential given that PP instances in our setting can differ in size.
Our ML model takes a graph representation of a single PP instance and returns an estimation of the probability of the optimal value of this instance to be negative.
The higher the estimated probability, the more likely the PP is to return a useful column.
We train the ML model with supervised learning.
We create the training set by collecting samples from numerous PP instances and labeling them by solving the instances with an exact algorithm.
We sum up the contributions of this paper in the following:
\begin{enumerate}
    \item We extend previous work on optimizing team formation and routing by incorporating ML into the branch-and-price algorithm.
    \item We define a novel partial column generation strategy that uses predictions to decide which PPs to solve at every iteration of the CG procedure.
    \item We propose a partially bipartite graph representation of the PPs as input for the GNN, we design a neural architecture tailored to this representation, and we validate its effectiveness with an ablation study.
    \item We verify the efficacy of the proposed technique by benchmarking its performance against other partial column generation strategies from the literature.
\end{enumerate}

\section{Literature Review}
In this section, we summarize the main publications related to this topic.
In Section \ref{sec:PCGliterature}, we gather the most relevant publications that propose and apply partial column generation strategies.
We compare the performance of our approach against these strategies, if applicable to our problem, in Section~\ref{sec:computationalStudy}.
In Section \ref{sec:ml4cg}, we review other publications that use ML techniques to accelerate CG. 

\subsection{Partial Column Generation} \label{sec:PCGliterature}
Partial column generation is a partial pricing technique to accelerate column generation.
The idea of partial column generation was first introduced by \citet{gamache1999column}.
In this publication, the authors fix the maximum number of PPs returning a column with a negative reduced cost in a single iteration.
At each CG iteration, they solve PPs until either the maximum number is hit or all PPs are solved.
The ordering of the PPs is arbitrarily fixed and it remains unchanged throughout the whole execution.
Each CG iteration starts solving the PP immediately following the last one solved in the previous iteration, in a round-robin fashion.
The authors determine the maximum number of PPs empirically through trial-and-error by testing different values.

Most publications applying partial column generation implement the strategy of \citet{gamache1999column} and empirically adjust the maximum number of PPs to best fit their specific problem.
The maximum number of PPs can vary from 1 (e.g. \citet{tilk2018combined}, \citet{basso2022distributed}) to any number smaller than the total number of PPs (e.g. \citet{maenhout2010branching}, \citet{rezanova2010train}, \citet{van2021column} \citet{breugem2022column}).
\citet{quesnel2020branch} propose an extension of this strategy including an additional termination criterion.
Specifically, they define a maximum number of PPs failing to return a negative column.
If this limit is hit and at least one negative column is found, the CG proceeds with a new iteration.
\citet{koutecka2025machine} apply a variation of the strategy of \citet{gamache1999column} in which they use an ML ranking model to sort the PPs instead of selecting them in a round-robin fashion. 
Their ML model is trained to rank the PPs at a specific CG iteration, thus prioritizing those expected to yield better columns.

\citet{rothenbacher2016branch} propose a different partial column generation strategy based on the history of the previous CG iterations.
In the first CG iteration, they solve all PPs while, in the following ones, they only solve the PPs that returned a negative column in the previous CG iteration. 
\citet{ruther2017integrated} propose a sorting of the PPs according to a problem-specific score that considers the history and dual values of the convexity constraint.
They dynamically adjust the maximum number of PPs according to the instance size.






\subsection{Machine Learning for Column Generation} \label{sec:ml4cg}
Using ML techniques to accelerate CG is a relatively new literature stream.
Several publications focus on reducing the size of the graph of PPs that constitute a (E)SPPRC, falling within the scope of heuristic pricing. 
For example, \citet{quesnel2022deep} use a deep neural network in the context of personalized crew rostering to select and remove unpromising nodes from the PP graph.
\citet{tahir2021improved} estimate the probability of the arcs and nodes to be part of a near-optimal solution of the PP with a convolutional neural network.
The authors remove arcs with low probability from the graph of the PP, and incorporate the probability as resource of the partial path, discarding those that do not seem promising.
\citet{morabit2023machine} compare a few ML models to reduce the size of the SPPRC graph for two different variations of VRP.
All the ML models of these publications focus either on the single arcs or nodes and have, therefore, very little awareness of the topology of the graph.
\citet{gerbaux2025machine} combine a greedy heuristic with GNN to identify arcs from the PP graph. 

When a PP constitutes a (E)SPPRC, the labeling algorithm used to solve it can produce multiple columns.
Usually, only the column with the most negative reduced cost is added to the RMP.
However, this does not always result in the fastest convergence.
\citet{morabit2021machine} and \citet{chi2022deep} develop a GNN model based on the bipartite graph representation by \citet{gasse2019exact} of the RMP to select the most promising columns to include. 

Some variations of VRP, such as the 2D-constrained VRP, might require heavy feasibility checks to verify the feasibility of the columns yielded by the PPs.
\citet{xia2024neural} propose an ML model to predict the feasibility of a column, reducing the number of necessary feasibility checks.

\citet{shen2024adaptive} use an ML to predict optimal dual solutions, which they feed to an adaptive stabilization technique to accelerate the convergence of CG to solve the graph coloring problem.

\subsection{Positioning of our Approach in the Literature}
The most closely related publication to ours is that of \citet{koutecka2025machine}, where the authors apply ML techniques to guide their PCG strategy.
The key difference between their approach and ours is that we focus on selecting which PPs to solve, rather than ranking them. 
To the best of our knowledge, our publication is the first one proposing a PCG strategy where ML is used to select which PPs to solve. 

\section{The Team Formation and Routing Problem}
The first publication optimizing the formation and routing of teams for airport baggage loading and unloading was \citet{dall2023formation}.
\citet{hagn2024branch} subsequently extended the problem by introducing stochastic travel times and proposed an adapted solution method. 
The goal of the planning problem is to form teams of skilled workers and route them across the apron to perform the loading and unloading tasks.
Each task can be carried out in different modes, each requiring a different composition of the team and affecting the execution time of the task.
The travel time between locations is stochastic.
The objective function pushes the tasks to terminate as soon as possible and reduces their risk of being overdue.

\subsection{Problem Definition}\label{sec:problemDescription}
The problem is defined on a discrete planning horizon with discrete time points $\tau\in\mathcal{T}$.
Each worker possesses a skill level $k \in \mathcal{K}$.
Skill levels are hierarchical, meaning that a worker with a certain skill level $k$ can operate any level $k' \le k$.
For each skill level $k$, parameter $\mathcal{N}_k$ represents the number of workers available with skill level $k'' \geq k$.
Each team must comply with at least one of the possible working profiles $q \in \mathcal{Q}$.
A working profile determines a feasible configuration of a team, hence a number of workers for each skill level.
A working profile $q$ is defined by the tuple $(\xi_{q,k})_{k \in\mathcal{K}}$ where $\xi_{q,k}$ represents the number of workers required with at least skill level $k$.
Each task $i$ has a set of compatible working profiles $\mathcal{Q}_i$.
Only a team that complies with one of the working profiles $q \in \mathcal{Q}_i$ can carry out task $i$.
For a task $i$, profile $q\in \mathcal{Q}_i$ determines its execution time $p_{i,q}$.
Working profiles that require more or more skilled workers are associated with shorter execution times for the same task.
Each task $i$ is associated with a time window $\left[\mathrm{ES}_i, \mathrm{LF}_i\right]$.
A task cannot begin before its earliest start time $\mathrm{ES}_i$.
A task can end beyond its latest finish time $\mathrm{LF}_i$, incurring a high penalty, but never after its extended latest finish time $\mathrm{LF}^\mathrm{e}_i \ge \mathrm{LF}_i$.
The travel times between locations constitute discrete random variables.
For each pair $(i,j)\in \mathcal{E}$, a non-negative finite support vector $B_{i,j} \subset \mathbb{N}$ with $|B_{i,j}| < \infty$ defines the possible travel times from location $i$ to location $j$.
The vector of stochastic travel times $t_{i,j}: B_{i,j} \rightarrow \mathbb{N}_{\geq 0}$ is given by $t_{i,j}(\omega)$ for all $\omega\in B_{i,j}$.
For a pair $(i,j)$, we define $\pi_{i,j}: B_{i,j} \rightarrow \left[ 0,1\right]$ as the probability $\mathbb{P}(t=t_{i,j}(\omega))$ for all $\omega\in B_{i,j}$.
Since we assume all events in $B_{i,j}$ to be independent, the following condition holds
\begin{equation*}
    \sum_{\omega \in B_{i,j}} \pi_{i,j}(\omega) = 1
\end{equation*}
To indicate the worst-case travel time, that is, the largest one, we use 
\begin{equation*}
    \omega_{\mathrm{max}} = \left(\max\{\omega: \omega\in B_{i,j}\}\right)_{(i,j)\in E}
\end{equation*}
A team route is a tuple $(r,q)$ where $q$ represents a working profile, and $r$ is a feasible route.
For a route $r$ executed by a team with profile $q$ and visiting the sequence of tasks $\mathcal{I}^r$, the following two conditions must hold for feasibility.
\begin{align}
    &\mathbb{P}(F_i^{r,q} \leq \mathrm{LF}_i) \geq \alpha & \forall i \in \mathcal{I}_q\label{chance_constraint},\\
    &\mathbb{P}(F_i^{r,q} > \mathrm{LF}_i^{\mathrm{e}}) = 0  & \forall i \in \mathcal{I}_q\label{finish_at_lf_v}.
\end{align}
where $F_i^{r,q}$ represents the finish time of task $i$ in the route and $\alpha \in \left[1,0\right]$ is the minimum service level. 
Constraint (\ref{chance_constraint}) prevents the probability of a task violating the latest finish time to exceed the minimum service level $\alpha$.
Constraint (\ref{finish_at_lf_v}) guarantees that the task does not terminate beyond the extended latest finish time $\mathrm{LF}^\text{e}_i$.
The expected cost of a route $r$ with profile $q$ is
\begin{equation*}
    \mathbb{E}(c^r) = \sum_{i \in I_r} w_i \mathbb{E}((F^{r,q}_i - \mathrm{EF}_i) + P_i(F^{r,q}_i))
\end{equation*}
where $F^{r,q}_i$ is the finish time of task $i$ in the route, $\mathrm{EF}_i = \mathrm{ES}_i + \min_{q \in \mathcal{Q}_i} \left\lbrace p_{i,q} \right\rbrace$ is the earliest finish time, and
\begin{equation*}
    P_i(F_i^{r,q}(\omega)) = \begin{cases}
    \begin{aligned}
        &(F_i^{r,q}(\omega) - \mathrm{LF}_i)^2 && \text{if} \ \ F_i^{r,q}(\omega) > \mathrm{LF}_i,\\
        &0 && \text{otherwise}
    \end{aligned}
    \end{cases}
\end{equation*}
is a quadratic penalty function.
Given pair of consecutive tasks $(i,j)$ carried out in a team route $(r,q)$, the distribution of the finish times of $j$ is given by
\begin{equation}
    \mathbb{P}(F^{r,q}_j = \tau) = \begin{cases}
        \begin{aligned}
            &\underset{z=0}{\overset{\mathrm{ES}_{j} + p_{j,q} }{\sum}} \mathbb{P}(F^{r,q}_{i} + t_{i,j}  + p_{j,q} = z) && \text{if} \ \ \tau = \mathrm{ES}_{j}  + p_{j,q} \\
            &\mathbb{P}(F^{r,q}_i + t_{i,j} + p_{j,q} = \tau) && \text{if} \ \ \tau > \mathrm{ES}_j  + p_{j,q} \\
            &0 && \text{otherwise}
        \end{aligned}
    \end{cases}\label{calcul_start_time_distr}
\end{equation}

\section{Branch-and-Price Solution Approach}
In this section, we describe the branch-and-price solution method used to solve the problem.
This algorithm corresponds to the basic version of the solver configuration proposed in \citet{hagn2024branch} with the addition of the branching strategy based on the finish times of the tasks. 
We present the ML acceleration in Section \ref{sec:methodology}

\subsection{Master Problem}
We now introduce the master problem, which is formulated as a set covering problem.
The set $\mathcal{R}$ contains all possible feasible team routes $(r,q)$.
For each team route $(r,q)$, a binary variable $\lambda_q^r$ indicates whether the team route is part of the solution or not.
Parameter $b_{k,\tau}^{r,q}$ indicates how many workers with at least skill level $k$ team route $(r,q)$ employs at time $\tau$ assuming that the worst-case travel time $\omega_\mathrm{max}$ occurs.
This number is constant between the leave time $\mathrm{tl}^{r,q}$ of the team route from the depot and its return time $\mathrm{tr}^{r,q}$ to the depot and is equal to $0$ elsewhere.
The formulation of the master problem is:
\begin{align}
    \min & \sum_{(r,q)\in \mathcal{R}} \mathbb{E}(c^r)\lambda^r_q \label{model:obj_aggregated_mp}\\
    \textrm{s.t.} \quad & \underset{(r,q)\in\mathcal{R}: \ i\in\mathcal{I}^r}{\sum} \lambda^r_q \geq 1 \quad \forall i\in\mathcal{I}\label{model:tasksdone_aggregated_mp}\\
    & \underset{(r,q)\in\mathcal{R}}{\sum}b_{k,\tau}^{r,q}\lambda^r_q\leq N_k \quad \forall k\in\mathcal{K},\ \forall \tau\in \mathcal{T} \label{model:workerconstr_aggregated_mp}\\
    & \lambda^r_q\in \left\lbrace 0,1 \right\rbrace \quad \forall (r,q)\in\mathcal{R}\label{model:lambda_binary_aggregated_mp}
\end{align}
The reduced master problem has the same formulation as the MP but considers a subset of variables $\bar{\mathcal{R}}\subset \mathcal{R}$ and relaxes the integrality constraint.

\subsection{Pricing Problems}
The pricing problems are responsible for generating new feasible team routes.
For each working profile $q \in \mathcal{Q}$, we have a distinct pricing problem.
The objective function of a PP, which corresponds to the reduced cost of a column representing a team route $(r,q)$, is
\begin{equation*}
    \mathbb{E}(c^r) - \underset{i\in\mathcal{I}^r}{\sum} \mu_i - \underset{k\in\mathcal{K}}{\sum} \ \underset{\tau=\mathrm{tl}^{r,q}}{\overset{\mathrm{tr}^{r,q}}{\sum}}\delta_{k,\tau}b_{k,\tau}^{r,q}\label{reduced_cost_formula_amp}
\end{equation*}
where $\mu_i$ are the dual variables relative to Constraint (\ref{model:tasksdone_aggregated_mp}) and $\delta_{k,\tau}$ are the dual variables relative to Constraint (\ref{model:workerconstr_aggregated_mp}).
Each PP is defined on an independent pricing network and constitutes an ESPPRC.
The pricing network for a profile $q \in \mathcal{Q}$ is a directed graph $\mathcal{G}_q = (\mathcal{V}_q, \mathcal{A}_q)$.
The set $\mathcal{V}_q$ contains a node for each task $i \in \mathcal{I}_q$ plus a source node $o$ and a sink node $o'$ representing departure from and return to the depot, respectively.
The set $\mathcal{A}_q$ contains arcs from the source node $o$ to all transportation nodes and from all transportation nodes to the sink node $o'$ as well as an arc for each ordered pair of tasks $(i,j) \in \mathcal{I}_q \times \mathcal{I}_q$.
From those, we remove any arc that cannot comply with the feasibility conditions of a team route from the set of arcs $\mathcal{A}_q$. 
If completing task $i$ early enough to reach and perform task $j$ within its latest finish time $LF_j$ with at least probability $\alpha$ is not possible, the feasibility constraint (\ref{chance_constraint}) cannot hold.
Similarly, the feasibility constraint (\ref{finish_at_lf_v}) cannot hold if task $i$ cannot terminate early enough to travel to task $j$ and complete the latter within its extended finish time $LF^\mathrm{e}_i$ in case the worst-case travel time occurs. 
Furthermore, we can remove all arcs $(i,j)$ for which the worst-case travel time from task $i$ to task $j$ is not smaller than worst-case travel time from $i$ to the depot $o$ plus the worst-case travel time from the depot $o$ to $j$.
Formally, we can remove arc $(i,j)$ from $\mathcal{A}_q$ if
\begin{equation}
    \mathrm{ES}_i + p_{i,q} + t_{i,j}^{\alpha} > \mathrm{LF}_j - p_{j,q}
\end{equation}
or 
\begin{equation}
    \mathrm{ES}_i + p_{i,q} + t_{i,j}(\omega_{\max}) > \mathrm{LF}_j - p_{j,q}
\end{equation}
or 
\begin{equation}
    \mathrm{ES}_j - \mathrm{LF}_i^{\mathrm{e}} \geq t_{i,o}(\omega_\mathrm{max}) + t_{o,j}(\omega_\mathrm{max})
\end{equation}
holds, being
\begin{equation*}
    t_{i,j}^{\alpha} = \min\left\{t^*: \ \mathbb{P}(t_{i,j} \leq t^*) \geq \alpha \right\}
\end{equation*}
the largest $\alpha$-quantile of the distribution of $t_{i,j}$.

We call a path from the source node $o$ to the sink node $o'$ on a specific pricing network a full path.
A full path alone is not sufficient to represent a team route, as the pricing networks lack temporal information.
For a full path visiting a sequence of tasks $\mathcal{I}_r$ on pricing network $\mathcal{G}_q$, a leave time $\mathrm{tl}^{r,q}$ determines the expected finish times of the visited tasks $i \in \mathcal{I}_r$.
The combination of a full path on pricing network $\mathcal{G}_q$ and a leave time $\mathrm{tl}^{r,q}$ represents a feasible team route $(r,q)$ if the feasibility conditions (\ref{chance_constraint}) and (\ref{finish_at_lf_v}) are met.
Given a partial path and a leave time, the expected cost of an arc along the path is
\begin{equation*}
    W_{i,j}\left(F_i^P(\omega_{\mathrm{max}})\right) = w_{j} \cdot \mathbb{E}(F_{j} + P_{j}(F_{j})) -\mu_{j} -  \underset{\tau=F_{i}^{P}(\omega_{\mathrm{max}})+1}{\overset{F_{j}^P(\omega_{\mathrm{max}})}{\sum}}  \ \underset{k\in\mathcal{K}}{\sum} \delta_{k,\tau} \xi_{q,k}
\end{equation*}
The sum of the expected costs of the arcs along a feasible full path, given a leave time $\mathrm{tl}$, corresponds to the reduced cost of the column.

\subsection{Labeling Algorithm for the Pricing Problems}
To find elementary shortest paths in the pricing networks, we use the labeling algorithm for ESPPRC described in \citet{errico2018vehicle}.
The resources ensure the elementarity of the path and the feasibility of the corresponding team route.
The labeling algorithm begins by calculating the initial labels, that is, one for each possible discrete start time of each task in the pricing network.
The labels are extended along the arcs to generate new partial paths as long as the resource are not depleted or the sink node is reached.
A dominance rule prunes labels that cannot lead to an optimal solution to the PP.

\subsection{Branching Strategies} \label{sec:branchingStrategies}
To find integer solutions, it is necessary to branch.
Three branching rules are proposed.

\paragraph{First Branching Rule}
The first branching rule consists of branching on the worst-case finish time of a task.
Given an optimal fractional solution to the RMP, let $i$ be a task carried out in two different team routes $(r',q')$ and $(r'',q'')$ such that $F^{r',q'}_i \neq F^{r'',q''}_i$.
We create two branches by imposing $F^{r,q}_i \leq \tau$ in one child node and $F^{r,q}_i > \tau$ in the other child node. 
We select $\tau$ so that none of the solution spaces of the children node are equal to the solution space of the parent node.

\paragraph{Second Branching Rule}
In case the first branching strategy is not applicable, we resort to the second branching rule, that is, branching on the number of tours at a given time.
For a given team route $(r,q)$, we define the parameter $g_\tau^r=1$ for $\mathrm{tl}^{r,q}\leq\tau\leq \mathrm{tr}^{r,q}$, and $g_\tau^r=0$ otherwise.
Given an optimal solution $\lambda^*$ to the RMP, let $\tau$ be an instant such that $l_\tau = \sum_{(r,q) \in \mathcal{R}}g_\tau^r \lambda_q^r$ is fractional.
We create two branches by imposing
\begin{equation*}
    \underset{(r,q)\in\mathcal{R}}{\sum}g_{\tau^{*}}^r\lambda^r_q \leq \lfloor l_{\tau^{*}}\rfloor \quad \text{or} \quad
    \underset{(r,q)\in\mathcal{R}}{\sum}g_{\tau^{*}}^r\lambda^r_q \geq \lceil l_{\tau^{*}}\rceil.
\end{equation*}
in the MP of the children nodes.
This branching rule introduces new dual variables $\rho_\tau$ and $\gamma_\tau$
for each time $\tau$ where the branching constraints apply.
In such cases, the values of the dual variables $\rho_\tau$ and $\gamma_\tau$ must be subtracted to the expected cost of an arc.

\paragraph{Third Branching Rule}
In case the first and second branching strategies are not applicable, we resort to branching on variables.
We select a fractional variable $\lambda^{r^*}_{q^*}$ and fix its value to $1$ in a branching child and to $0$ in the other one.
In the pricing problem, we ensure that the column is not generated again by introducing a resource that decreases whenever the label is extended without diverging from the partial path corresponding to the column.

\subsection{Feasibility Check}
The RMP considers workers on an aggregated level, therefore, integer solutions are not guaranteed to be operationally feasible.
An operationally infeasible solution is a feasible integer solution to the RMP for which there exists no feasible assignment of single workers to the teams.
To verify the operational feasibility of an integer solution, we solve the feasibility problem proposed by \citet{dall2023formation}.
We report the feasibility check in \ref{app:feasibilityCheck}. 
If an integer solution $\bar{\lambda}$ does not pass the feasibility check, we prohibit $\bar{\lambda}$ with the following cut in the master problem
\begin{equation}
    \sum_{(r,q) \in \bar{\mathcal{R}}} \lambda^r_q \leq |\bar{\mathcal{R}}| - 1
\end{equation}
where $\bar{\mathcal{R}} = \left\{ (r,q) \in \mathcal{R} : \bar{\lambda}^r_q = 1\right\}$ is the set of team routes that make up an integer solution $\bar{\lambda}$ to the RMP.
Applying this cut requires forbidding the repeated generation of the tours in $\bar{\mathcal{R}}$, which we achieve applying the same approach used for the second branching strategy in Section \ref{sec:branchingStrategies}.

\subsection{Early Termination Heuristic} \label{sec:earlyTerminationHeuristic}
We can define early termination criteria, such as a time limit, and stop the execution of the algorithm before it finds an optimal solution.
In this case, we try to quickly compute a feasible solution using the column that the CG has generated so far.
To do so, we combine the RMP and the feasibility problem and impose the integrality constraint on the variables $\lambda$.

\section{Partial Column Generation with GNNs}\label{sec:methodology}
In this section, we describe our PCG approach in detail.
The approach relies on predicting the sign of the optimal solution of the PPs to select which of them to solve at each CG iteration.
In the reminder of the paper, we refer to a column corresponding to a variable with a negative reduced cost, hence a column with the potential to improve the current solution to the RMP, as ``negative column''.
Similarly, we use the term ``non-negative column'' when the reduced cost of the variable is non-negative.
Section \ref{sec:pcgWithPred} shows how we adapt the CG procedure to incorporate predictions.
In Section \ref{sec:gnnForPP}, we explain how we use graph neural networks to make predictions and how we train our ML model.

\subsection{Partial Column Generation with Predictions} \label{sec:pcgWithPred}
Let us briefly recap the workflow of CG with full column generation.
The algorithm begins by initializing the RMP with a small set of columns.
We proceed by solving the RMP to optimality and we obtain a solution that serves as current feasible - possibly optimal - solution to the MP.
Then we integrate the optimal dual solution of the RMP into the PPs and solve them all to optimality.
Some PPs might turn out to be infeasible and, therefore, return no column.
Each of the other PPs returns either a negative or a non-negative column.
If we obtain negative columns, we add them to the RMP and we start the following iteration of the CG by solving the RMP once again and repeating the previous steps.
Note that we discard all non-negative columns since they cannot contribute to improve the current MP solution.
If solving all PPs to optimality does not supply any negative columns, we have proven that no variable of the MP has potential to improve the current solution.
This means that the current RMP optimal solution is also optimal to the MP, which concludes the procedure.

Let us analyze a non-terminal iteration, that is, one where at least one PP returns a negative column.
In such an iteration, solving PPs that do not yield a negative column has no effect other than consuming computational resources.
In fact, these PPs either return no column or return non-negative columns, which are simply discarded.
If we knew in advance that a PP is not useful, we could simply skip it and this would have no impact on the behavior of the algorithm while sparing computational effort.
In general, it is not possible to determine with certainty whether a PP will return a negative column without solving it.
However, we can make a prediction on the sign of the optimal value and only solve the PPs that we expect to be useful.
We base our partial column generation strategy exactly on this idea.

\begin{figure}
    \centering
    \begin{tikzpicture}[
  node distance=2cm and 1.5cm,
  startstop/.style={rectangle, rounded corners, minimum width=3cm, minimum height=1cm,text centered, draw=black},
  process/.style={rectangle, minimum width=2.5cm, minimum height=1cm, text centered, draw=black},
  decision/.style={diamond, minimum width=2.5cm, minimum height=1cm, aspect=2, text centered, draw=black},
  arrow/.style={->, thick, >=stealth},
  font=\small
  ]
  
\node (start) [startstop] {Start};
\node (rmp) [process, below of=start] {Solve RMP};
\node (gnn1) [decision, below of=rmp, line width=2pt, xshift=-4.2cm] {\shortstack{PP 1\\useful?}};
\node (gnn2) [decision, below of=rmp, line width=2pt] {\shortstack{PP 2\\ useful?}};
\node (gnnn) [decision, below of=rmp, line width=2pt, xshift=4.2cm] {\shortstack{PP N\\ useful?}};
\node (dots_gnn) [text centered] at ($(gnn2)!0.5!(gnnn)$) {\textbf{\dots}}; 
\node (pp1) [process, below of=gnn1] {Solve PP 1};
\node (pp2) [process, below of=gnn2] {Solve PP 2};
\node (ppn) [process, below of=gnnn] {Solve PP N};
\node (dots_pp) [text centered] at ($(pp2)!0.5!(ppn)$) {\textbf{\dots}}; 
\node (check1) [decision, below of=pp2, yshift=-0.5cm] {Negative Columns?};
\node (update) [process, right of=check1, xshift=4cm] {\shortstack{Add Column(s)\\ to RMP}};
\node (remaining) [process, below of=check1, line width=2pt, yshift=-0.5cm] {Solve Remaining PPs};
\node (check2) [decision, below of=remaining, line width=2pt, yshift=-0.5cm] {Negative Columns?};
\node (end) [startstop, below of=check2, yshift=-0.5cm] {End};

\draw [arrow] (start) -- (rmp);
\draw [arrow] (rmp) -- (gnn1);
\draw [arrow] (rmp) -- (gnn2);
\draw [arrow] (rmp) -- (gnnn);
\draw [arrow] (gnn1) -- (pp1) node[near start, right] {yes};
\draw [arrow] (gnn2) -- (pp2) node[near start, right] {yes};
\draw [arrow] (gnnn) -- (ppn) node[near start, right] {yes};
\draw [arrow] (pp1) -- (check1);
\draw [arrow] (pp2) -- (check1);
\draw [arrow] (ppn) -- (check1);
\draw [arrow] (check1.east) -- (update.west) node[pos=0.15, above] {yes};
\draw [arrow] (check1) -- (remaining) node[near start, right] {no};
\draw [arrow] (remaining) -- (check2);
\draw [arrow] (check2.east) -| (update.south) node[pos=0.05, above] {yes};
\draw [arrow] (update.north) |- (rmp.east);
\draw [arrow] (check2.south) -- (end.north) node[near start, right] {no};

\newdimen\Xgnn
\newdimen\Ygnn
\pgfextractx{\Xgnn}{\pgfpointanchor{gnn1}{west}}
\pgfextracty{\Ygnn}{\pgfpointanchor{gnn1}{west}}
\coordinate (kink) at (\dimexpr\Xgnn-0.8cm, \Ygnn);

\draw [arrow] (gnn1.west) |- (kink) node[near start, above left] {no} |- (check1.west) ;

\end{tikzpicture}
    \caption{Modified workflow of CG to include predictions. The additional components that are not part of the traditional CG workflow are highlighted with a thick border}
    \label{fig:workflowCG}
\end{figure}
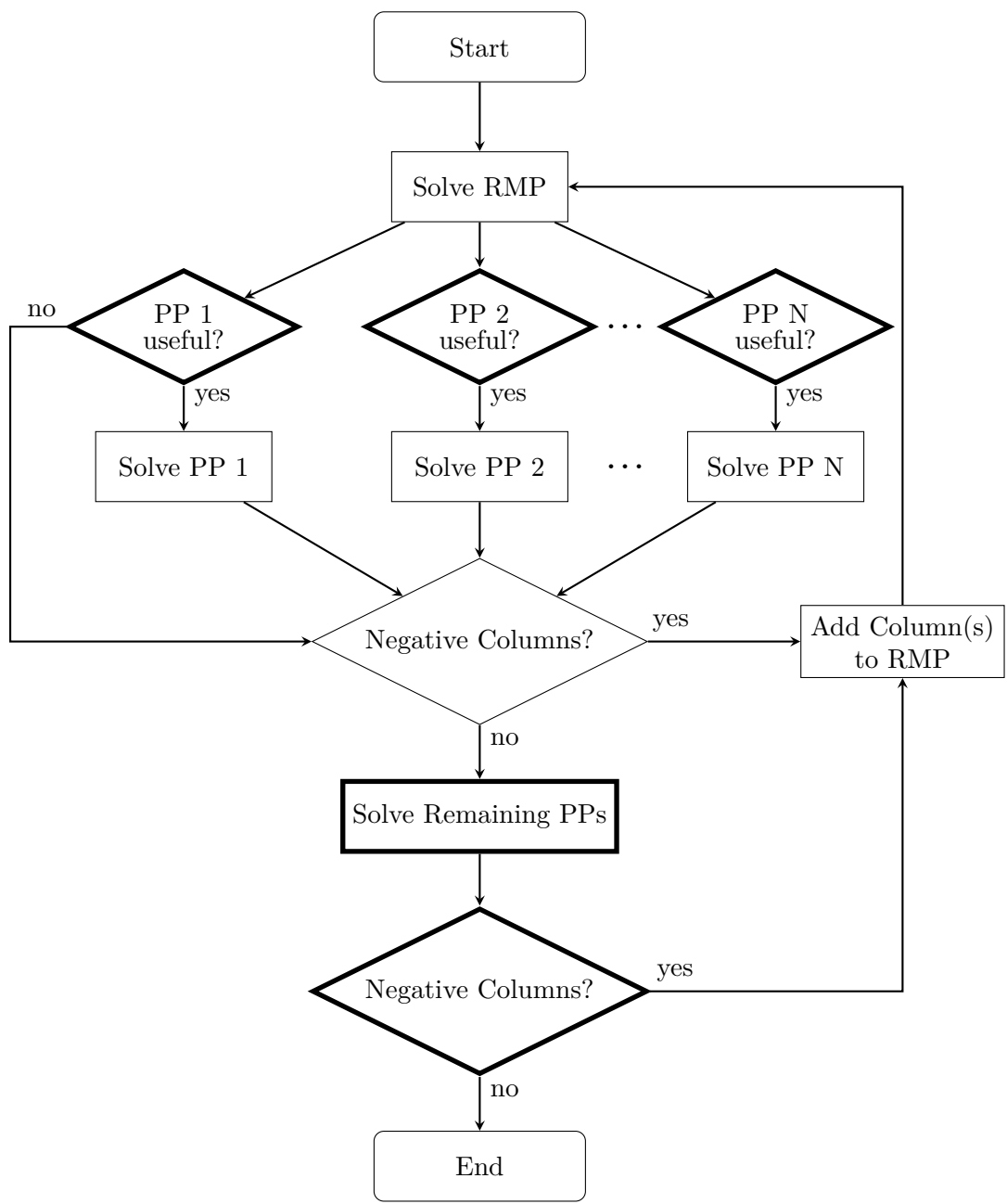

To realize our PCG strategy, we modify the workflow of the CG procedure as follows.
We introduce a new component, a predictor, that we invoke after solving the RMP and before solving the PPs.
This predictor estimates whether the optimal value of each specific PP instance is negative and, therefore, whether the PP will produce a negative column.
We then solve only the PPs that the predictor expects to yield a negative column.
If the selected PPs produce one or more negative columns, we add them to the RMP and proceed with the following iteration.
If the selected PPs do not return any negative columns or if the predictor does not select any PP, we may be in a terminal iteration.
To confirm that there are no more negative columns, we need to solve all remaining PPs.
In fact, the predictions might not all be correct and we might have skipped a PP with a negative column.
If we obtain some new ones, we add them to the RMP and resume the loop by solving it again.
Otherwise, we have found an optimal solution and the algorithm terminates.
Figure \ref{fig:workflowCG} illustrates our CG workflow modified to include the predictions.


\subsection{Graph Neural Networks for Pricing Problem Prediction} \label{sec:gnnForPP}
In this section, we describe how we make our predictions on the sign of the optimal solution of the PPs using ML.
Specifically, we detail how we represent a PP instance as a partially bipartite graph that we feed to our GNN in Section \ref{sec:inputGraph}, we describe the neural architecture of our ML model in Section \ref{sec:neuralArchitecture}, and we outline the algorithm that we use to train our model in Section \ref{sec:TrainingAlgorithm}.

\subsubsection{Partially Bipartite Input Graph}\label{sec:inputGraph}
Graph neural networks take graphs as input, therefore, we need to represent the single PP instances as graphs.
The single representation serves as input graph that we can feed to our GNN and make the prediction.
The graph can represent the information of the PP instance explicitly as features of nodes or arcs, or implicitly in the topology of the graph.
Since each PP instance is specific to a working profile $q \in \mathcal{Q}$, the corresponding input graph $\hat{\mathcal{G}}_q = (\hat{\mathcal{V}}_q, \hat{\mathcal{A}}_q)$ is also dependent on $q$. 
Our input graph is partially bipartite.


\begin{definition}
We define a \emph{partially bipartite graph} $\mathcal{G} = (\mathcal{V}, \mathcal{A})$ as a graph composed of two disjoint sets of nodes, $\mathcal{V}^S$ and $\mathcal{V}^T$, a set of undirected edges $\mathcal{A}^S$, and a set of directed arcs $\mathcal{A}^T$.
The overall node and arc sets of the partially bipartite graph are defined as:
\[
\mathcal{V} = \mathcal{V}^S \cup \mathcal{V}^T \quad \text{and} \quad \mathcal{A} = \mathcal{A}^S \cup \mathcal{A}^T,
\]
where
\[
\mathcal{V}^S \cap \mathcal{V}^T = \emptyset \quad \text{and} \quad \mathcal{A}^S \cap \mathcal{A}^T = \emptyset.
\]
Edges in $\mathcal{A}^S$ only connect nodes from different node sets, i.e.:
\[
\forall \{u, v\} \in \mathcal{A}^S: \quad (u \in \mathcal{V}^S \land v \in \mathcal{V}^T) \lor (u \in \mathcal{V}^T \land v \in \mathcal{V}^S),
\]
while arcs in $\mathcal{A}^T$ only connect nodes within $\mathcal{V}^T$, i.e.,
\[
\forall (u, v) \in \mathcal{A}^T: \quad u \in \mathcal{V}^T \land v \in \mathcal{V}^T.
\]
It follows that subgraph $\mathcal{G}^T = (\mathcal{V}^T, \mathcal{A}^T)$ forms an arbitrary directed graph, while subgraph $\mathcal{G}^S = (\mathcal{V}^S \cup \mathcal{V}^T, \mathcal{A}^S)$ forms an undirected bipartite graph.
\end{definition}

Given our partially bipartite input graph $\hat{\mathcal{G}}_q = (\hat{\mathcal{V}}_q, \hat{\mathcal{A}}_q)$, we refer to its arbitrary directed subgraph $\hat{\mathcal{G}}_q^T=(\hat{\mathcal{V}}_q^T, \hat{\mathcal{A}}_q^T)$ as the transportation graph, to its nodes in $\hat{\mathcal{V}}_q^S$ as supplementary nodes, and to its edges in $\hat{\mathcal{A}}_q^S$ as supplementary edges.
The transportation graph derives directly from the pricing network and it can easily incorporate the instance parameters that are defined on the set of nodes or on the set of arcs of a pricing network.
The transportation graph $\hat{\mathcal{G}}_q^T=(\hat{\mathcal{V}}_q^T, \hat{\mathcal{A}}_q^T)$ for working profile $q$ includes all nodes and arcs of the pricing network $\mathcal{G}_q$, excluding the source node $o$ and the sink node $o'$. 
Formally, the set of transportation nodes is $\hat{\mathcal{V}}_q^T = \left\{ i \mid i \in \mathcal{I}_q \right\}$ and the set of transportation arcs is $\hat{\mathcal{A}}_q^T =  \left\{(i,j) \mid (i,j) \in \mathcal{A}_q \land i \neq o \land j \neq o'\right\}$.
Since a pricing network does not account for temporal information, we use the supplementary nodes and edges to incorporate time-dependent parameter.
The set of supplementary nodes $\hat{\mathcal{V}}_q^S$ contains a node for each point in time $\tau \in \mathcal{T}$.
We use these nodes to incorporate the instance parameters of the PP problems that are defined on the set $\mathcal{T}$.
Each supplementary node $\tau \in \hat{\mathcal{V}}_q^S$ is connected to each transportation node $i \in \hat{\mathcal{V}}_q^T$ by a supplementary edge $(i, \tau) \in \mathcal{A}_q^S = \hat{\mathcal{V}}_q^T \times \hat{\mathcal{V}}_q^S$.
In the supplementary edges, we incorporate temporal information on the tasks derived from the time windows.
Figure \ref{fig:inputGraph} shows the structure of the partially bipartite input graph on an example.
In the remainder of this section, we describe in detail how we calculate the features of the partially bipartite input graph to obtain a unique representation of the PP instance.

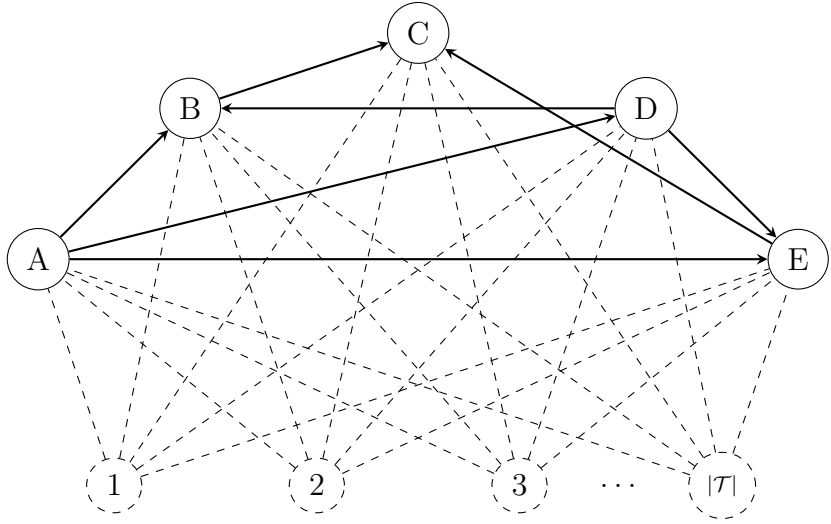
\begin{figure}
    \centering
    \begin{tikzpicture}[
    transArc/.style={thick,->, >=stealth},
    suppEdge/.style={-,dashed},
    transNode/.style={draw, circle},
    suppNode/.style={draw, circle, dashed},
    hiddenNode/.style={},
    ]
    
    \node[transNode] (A) at (-1,0) {A};
    \node[transNode] (C) at (4,3) {C};
    \node[transNode] (E) at (9,0) {E};
    \node[transNode] (B) at (1,2) {B};
    \node[transNode] (D) at (7,2) {D};
    
    \node[suppNode] (F) at (0,-3) {1};
    \node[suppNode] (G) at (2.66666,-3) {2};
    \node[suppNode] (H) at (5.33333,-3) {3};
    \node[suppNode] (I) at (8,-3) {\scriptsize{$|\mathcal{T}|$}};

    \node[hiddenNode] (L) at (6.66666, -3) {\dots};

    \draw [transArc] (A)  -- (B);
    \draw [transArc] (A)  -- (D);
    \draw [transArc] (A)  -- (E);
    \draw [transArc] (B)  -- (C);
    \draw [transArc] (D)  -- (B);
    \draw [transArc] (D)  -- (E);
    \draw [transArc] (E)  -- (C);
    
    \draw [suppEdge] (F)  -- (A);
    \draw [suppEdge] (G)  -- (A);
    \draw [suppEdge] (H)  -- (A);
    \draw [suppEdge] (I)  -- (A);
    \draw [suppEdge] (F)  -- (D);
    \draw [suppEdge] (G)  -- (D);
    \draw [suppEdge] (H)  -- (D);
    \draw [suppEdge] (I)  -- (D);
    \draw [suppEdge] (F)  -- (E);
    \draw [suppEdge] (G)  -- (E);
    \draw [suppEdge] (H)  -- (E);
    \draw [suppEdge] (I)  -- (E);
    \draw [suppEdge] (F)  -- (B);
    \draw [suppEdge] (G)  -- (B);
    \draw [suppEdge] (H)  -- (B);
    \draw [suppEdge] (I)  -- (B);
    \draw [suppEdge] (F)  -- (C);
    \draw [suppEdge] (G)  -- (C);
    \draw [suppEdge] (H)  -- (C);
    \draw [suppEdge] (I)  -- (C);

\end{tikzpicture}
    \caption{Illustration of the structure of the partially bipartite input graph. The nodes and arcs of the transportation graph are represented with solid lines, while the supplementary nodes and edges are represented with dashed lines.}
    \label{fig:inputGraph}
\end{figure}

\paragraph{Vectorial Representation of the Stochastic Travel Times}
In our application problem, the travel times between locations consist of discrete stochastic variables $t_{i,j}(\omega)$.
For each pair of locations $(i,j)$, the set $B_{i,j}$ defines the possible travel times from $i$ to $j$.
The cardinality $|B_{i,j}|$ can vary between different pairs of locations.
However, empirical observations indicate that this number remains relatively small in practice.
To incorporate travel time information as features in our input graph, we represent the stochastic travel times between two locations as a fixed-length vector.
We fix the length of this vector to $2M$, where $M$ is at least as large as the maximum observed cardinality of the travel time sets $B_{ij}$ in all instances.
Given a location pair $(i,j)$, the corresponding representation vector of stochastic travel times contains all possible travel times $t_{i,j}(\omega)$ as well as the associated probabilities $\pi_{i,j}(\omega)$ for all $\omega \in B_{i,j}$ .
In cases where $|B_{ij}|$ is smaller than $M$, we apply a padding strategy to maintain a consistent vector length across all sets.
Formally, the representation vector of the stochastic travel times $\hat{d}_{i,j}$ from the location $i$ to the location $j$ is
\begin{equation}
    \hat{d}_{i,j} =  \left[ t_{i,j}(\omega) \right]_{\omega \in B_{i,j}} \, \Vert \, \left[ t_{i,j}(\omega_\mathrm{max}) \right]_{M - |B_{i,j}|} \, \Vert \, \left[ \pi_{i,j}(\omega) \right]_{\omega \in B_{i,j}} \, \Vert \, \left[ 0\right]_{M - |B_{i,j}|} 
\end{equation}
where the symbol $\Vert$ represents the concatenation operator.

\paragraph{Features of the Transportation Graph}
For every node $i \in \hat{\mathcal{V}}_q$, the corresponding feature vector includes parameters of the PP instance that are defined on the set of tasks $\mathcal{I}$.
The feature vector of the node corresponding to task $i$ contains the following scalar features: the weight $w_i$, the processing time $p_{i,q}$, the worst-case travel time from and to the depot $t_{o,i}(\omega_\mathrm{max})$, and the value of the dual variable $\mu_i$.
Additionally, the feature vector contains the representation vector of the stochastic travel times $\hat{d}_{i,o}$ between the location of $i$ and the location of the depot $o$.
Table \ref{tab:taskParameters} lists all the elements of the feature vector of the nodes $i$ in the transportation graph.
\begin{table}[h]
\centering
\label{tab:taskParameters}
\begin{tabular}{lllc}
\hline
Name & Symbol & Domain & Size \\ 
\hline
Weight &$w_i$ & $\mathbb{N}_{\geq 0}$  & 1\\
Processing time&$p_{i,q}$ & $\mathbb{N}_{>0}$ & 1 \\
Dual of Constraint \ref{model:tasksdone_aggregated_mp}&$\mu_i$ & $\mathbb{R}_{\geq 0}$  & 1\\
Max. travel time depot & $t_{o,i}(\omega_\mathrm{max})$ & $\mathbb{N}_{\geq 0}$  & 1\\
Flag non dominated formation & ndf & $\{0,1\}$  & 1\\ 
Stochastic travel time depot & $\hat{d}_{i,o}$ & $\mathbb{N}_{\geq 0}$ & $2M$ \\
\hline
\end{tabular}
\caption{Transportation-node parameters in simple bipartite graph approach}
\end{table}
For an arc $(i,j) \in \hat{\mathcal{A}}_q$, the feature vector consists of the maximum travel time and the representation vector of the stochastic travel from $i$ to $j$.
Table \ref{tab:arcParameters} lists the features of the arcs in the transportation graph. 
\begin{table}[h]
\centering
\label{tab:arcParameters}
\begin{tabular}{lllc}
\hline
Name & Symbol & Domain & Size \\
\hline
Max. travel time & $t_{i,j}(\omega_\mathrm{max})$ & $\mathbb{N}_{> 0}$ & 1 \\
Stoch. travel time & $\hat{d}_{i,j}$ & $\mathbb{N}_{\geq 0}$ & $2M$ \\ \hline
\end{tabular}
\caption{Transportation-edge parameters in simple bipartite graph approach}
\end{table}
Notably, the worst-case travel time between locations is redundantly present as a scalar feature as well as a part of the representation vector of the stochastic travel times in the feature vectors of the nodes and of the arcs of the transportation graph.
The reason for this is that the worst-case travel time plays a double role in our application problem: it participates in the feasibility of a team route (see Conditions (\ref{chance_constraint}) and (\ref{finish_at_lf_v})) as well as in the expected finish time of a task.
All other stochastic travel times are involved only in the calculation of the expected finish times of the tasks.
It should be noted that the time window parameters $\mathrm{ES}_i$, $\mathrm{LF}_i$, and $\mathrm{LF}_i^\mathrm{e}$ are not part of the features of the transportation nodes despite being defined on the set $\mathcal{I}$.
The reason for this is that these parameters refer to a point on the time horizon and their value has little inherent meaning.
This makes them difficult to interpret for the ML model and might hinder learning if used as raw data, even with normalization.
To avoid this, we represent them implicitly in the supplementary edges.

\paragraph{Features of the Supplementary Nodes}
The features of the supplementary nodes $\tau \in \hat{\mathcal{V}}_q^T$ represent temporal parameters of the PP instance that are defined on the set $\mathcal{T}$.
These parameters are the dual values $\delta_{k,\tau}$ from Constraint (\ref{model:workerconstr_aggregated_mp}) and the dual values $\rho_{\tau}$ and $\gamma_{\tau}$ from the branching constraints of the second branching rule from Section~\ref{sec:branchingStrategies}.
These values only play a role in the calculation of the expected cost of an arc in a pricing network.
We can rewrite the expected cost of an arc $(i,j)$ as
\begin{equation*}
    W_{i,j}\left(F_i^P(\omega_{\mathrm{max}})\right) = w_{j} \cdot \mathbb{E}(F_{j} + P_{j}(F_{j})) -\mu_{j} -  \underset{\tau=F_{i}^{P}(\omega_{\mathrm{max}})+1}{\overset{F_{j}^P(\omega_{\mathrm{max}})}{\sum}} \zeta_\tau
\end{equation*}
where $\zeta_\tau = \underset{k\in\mathcal{K}}{\sum} \delta_{k,\tau} \xi_{q,k} -\rho_\tau - \gamma_\tau.$
We can see the component $\zeta_\tau$ only depends on the time $\tau$.
Therefore, we do not have to keep the values $\delta_{k,\tau}$, $\rho_{\tau}$, and $\gamma_{\tau}$ separate, as we can aggregate them into $\zeta_\tau$ without loss of information.
For a supplementary node $\tau \in \hat{\mathcal{V}}_q^S$, the only feature is hence parameter $\zeta_\tau$. 

\paragraph{Features of the Supplementary Edges}
For a supplementary edge $(i, \tau) \in \mathcal{A}_q^S$, we consider the parameters that define the time window for task $i$ (see Section \ref{sec:problemDescription}), which are $\mathrm{ES}_i$, $\mathrm{LF}_i$, and $\mathrm{LF}_i^\mathrm{e}$ as well as the earliest finish time $\mathrm{EF}_i = \mathrm{ES}_i + \min_q \left\{p_{i,q}\right\}$, the profile-specific earliest finish time $\mathrm{EFQ}_{i,q} = \mathrm{ES} + p_{i,q}$, and the profile-specific latest start time $\mathrm{LSQ}_{i,q} = \mathrm{LF}^\mathrm{e}-p_{i,q}$.
We consider both the earliest finish time $\mathrm{EF}_i$ and the profile-specific earliest finish time $\mathrm{EFQ}_{i,q}$ because the first is relevant to the feasibility of a team route $(r,q)$ while the second is relevant to its cost.
We do not include the values of these parameters directly as features of a supplementary edge $(i, \tau)$ but we use binary values indicating whether the specific parameter of task $i$ is greater or equal to $\tau$, instead.
This avoids explicitly dealing with the numeric values of the time windows, which carry little meaning and might hinder learning.

\subsubsection{Neural Architecture} \label{sec:neuralArchitecture}
The goal of our GNN is to perform binary classification, and hence to assign the partially bipartite input graph representing a PP instance to one of the two possible classes, namely the ``negative optimal solution'' class and ``non-negative optimal solution'' class.
The GNN transforms the data structured in the input graph into a real-valued scalar $\hat{y} \in \left[0, 1\right]$.
The value of $\hat{y}$ can be interpreted as a probability of a PP instance to belong to one of the two classes.
The architecture of our GNN consists of multiple layers.
At each layer, the input graph or part of it undergoes a transformation that affects either the vectorial representation of nodes or arcs, or the topology of the graph.
In the following, we describe the layers that we use in our architecture.

\paragraph{Embedding Layer}
The first layer of our architecture is the embedding layer.
To calculate the embeddings, we use multilayer perceptrons (MLPs) with one hidden layer and rectified linear units (ReLUs) as activation functions.
For each type of nodes and arcs, we have a different MLP.
This means that the embedding layer is made up of $4$ independent MLPs, one for the transportation nodes, one for the transportation arcs, one for the supplementary nodes, and one for the supplementary edges.
Note that the structure of our input graph does not take any ordering of the supplementary nodes into consideration.
However, this information is crucial for taking into account the sequentiality of the time points $\tau \in \mathcal{T}$, with which the supplementary nodes are associated.
To overcome this, we add a positional encoding \citep{vaswani2017attention} to the embedding of the supplementary nodes.
We design the embedding layer such that all obtained embeddings have the same dimensionality, as this simplifies the architecture of the other layers.

\paragraph{Convolutional Layers}
We use convolutional layers to pass messages between neighboring transportation nodes.
Our convolutional layers are based on the graph isomorphism network with edge features (GINE) presented in \citet{hu2019strategies}.
We distinguish between in-convolutional layers, where messages flow in the direction of the arcs, and out-convolutional layers, where messages flow in the opposite direction of the arcs.
The formulation of our in-convolutional layer is
\begin{align}
    \mathbf{h}_i^{(l+1)} = \mathrm{MPL}\left( \left( 1 + \epsilon \right) \mathbf{h}_i^{(l)} + \sum_{j \in \mathcal{N}^-(i)} \mathrm{ReLU}\left( \mathbf{h}_j^{(l)} + \mathbf{e}_{j,i}\right)\right) & \quad\forall i \in \hat{\mathcal{V}}^T_q
\end{align}
where $\mathbf{h}_i^{(l)}$ and $\mathbf{h}_i^{(l+1)}$ are the representation of node $i \in \hat{\mathcal{V}}^T_q$ before and after the application of the layer $l$, respectively.
Furthermore, set $\mathcal{N}^-(i) = \left\{j \mid (j,i) \in \hat{\mathcal{A}}^T_q \right\}$ is the in-neighborhood of node $i$, vector $\mathbf{e}_{j,i}$ is the representation of arc $(j,i) \in \hat{\mathcal{A}}^T_q$ and $\epsilon$ is a learnable parameter.
The formulation of the out-convolutional layer is equivalent but defined on the out-neighborhood of node $i$
$\mathcal{N}^+(i) = \left\{j \mid (i,j) \in \hat{\mathcal{A}}^T_q \right\}$.

\paragraph{Bipartite Convolutional Layers}
We use our bipartite convolutional layers to pass messages between the transportation and supplementary nodes.
Depending on whether we are sending messages from transportation nodes to supplementary node or vice versa, we have a different function:
\begin{align}
    \mathbf{h}_\tau^{(l+1)} = \mathrm{MLP}\left( \left( 1 + \epsilon \right) \mathbf{h}_\tau^{(l)} + \sum_{i : (i,\tau) \in \hat{\mathcal{A}}^S_q} \mathrm{ReLU}\left( \mathbf{h}_i^{(l)} + \mathbf{e}_{i, \tau}\right)\right) & \quad \forall \tau \in \hat{\mathcal{V}}^S_q
\end{align}
and
\begin{align}
    \mathbf{h}_i^{(l+1)} = \mathrm{MPL}\left( \left( 1 + \epsilon \right) \mathbf{h}_i^{(l)} + \sum_{i : (i,\tau) \in \hat{\mathcal{A}}^S_q} \mathrm{ReLU}\left( \mathbf{h}_\tau^{(l)} + \mathbf{e}_{i, \tau}\right)\right) & \quad \forall i \in \hat{\mathcal{V}}^T_q
\end{align}
where $\mathbf{e}_{i, \tau}$ is the representation of edge $(i,\tau) \in \hat{\mathcal{A}}^S_q$ and $\mathbf{h}_\tau^{(l)}$ and $\mathbf{h}_\tau^{(l+1)}$ are the representation of node $\tau \in \hat{\mathcal{V}}^S_q$ before and after the application of the layer $l$, respectively.

\paragraph{Pooling and Final Layer}
We use the pooling layer to aggregate the information of the nodes of the transportation graph and obtain a 1-dimensional representation of the graph.
Note that we disregard the transportation arcs and the supplementary nodes and edges from now on as we assume that the information is now integrated in the representation of the transportation nodes.
The final layer applies an MLP to the output of the pooling layer, reducing the dimensionality to a scalar, and applies a sigmoid function to obtain a value between $0$ and $1$.
Formally
\begin{equation}
    \hat{y} = \sigma \left( \mathrm{MLP} \left( \sum_{i \in \hat{\mathcal{V}}^T_q} h^{(l_\mathrm{LAST})} \right)\right).
\end{equation}

Figure \ref{fig:neuralNetwork} depicts the architecture of our GNN.
Each column represents the evolution of the feature matrix of each type of component in the input graph, i.e., the transportation nodes $i$, the transportation arcs $(i,j)$, the supplementary nodes $\tau$, and the supplementary edges $(i,\tau)$.
Each row corresponds to one of the layers described above.
The black arrows represent the flow of information between the different layers.
The gray arrows and the blocks below them simplify the visualization and do not represent any transformation.
When a block becomes absent in a level, it means that it is no longer necessary and we discard it.
The pooling and final layers, which are not depicted in the figure, receive the output of the last out-convolution as input.
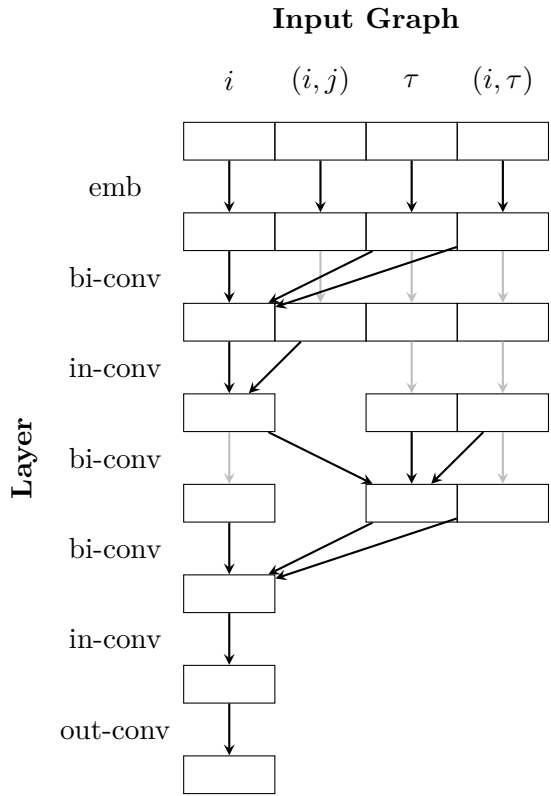
\begin{figure}[h]
    \centering
    \begin{tikzpicture}[
  node distance=1.3cm and 1.5cm,
  box/.style={draw, minimum width=1.2cm, minimum height=0.5cm, font=\tiny},
  arrow2/.style={->, thick, >=stealth},
  arrow3/.style={->, thick, lightgray, >=stealth},
  labelstyle/.style={font=\small}
  ]

  \node[labelstyle] at (1.8,12.4) {\textbf{Input Graph}};
  \node[labelstyle, rotate=90] at (-2.7,6.6) {\textbf{Layer}};

  \node[labelstyle] at (-1.5,10.2) {emb};
  \node[labelstyle] at (-1.5,9.0) {bi-conv};
  \node[labelstyle] at (-1.5,7.8) {in-conv};
  \node[labelstyle] at (-1.5,6.6) {bi-conv};
  \node[labelstyle] at (-1.5,5.4) {bi-conv};
  \node[labelstyle] at (-1.5,4.2) {in-conv};
  \node[labelstyle] at (-1.5,3.0) {out-conv};

  \node[labelstyle] at (0,11.6) {$i$};
  \node[labelstyle] at (1.2,11.6) {$(i,j)$};
  \node[labelstyle] at (2.4,11.6) {$\tau$};
  \node[labelstyle] at (3.6,11.6) {$(i,\tau)$};

  \node[box] (l00) at (0,10.8) {};
  \node[box] (l01) at (1.2,10.8) {};
  \node[box] (l02) at (2.4,10.8) {};
  \node[box] (l03) at (3.6,10.8) {};

  \node[box] (l10) at (0,9.6) {};
  \node[box] (l11) at (1.2,9.6) {};
  \node[box] (l12) at (2.4,9.6) {};
  \node[box] (l13) at (3.6,9.6) {};

  \node[box] (l20) at (0,8.4) {};
  \node[box] (l21) at (1.2,8.4) {};
  \node[box] (l22) at (2.4,8.4) {};
  \node[box] (l23) at (3.6,8.4) {};

  \node[box] (l30) at (0,7.2) {};
  \node[box, draw=none] (l31) at (1.2,7.2) {};
  \node[box] (l32) at (2.4,7.2) {};
  \node[box] (l33) at (3.6,7.2) {};

  \node[box] (l40) at (0,6.0) {};
  \node[box, draw=none] (l41) at (1.2,6.0) {};
  \node[box] (l42) at (2.4,6.0) {};
  \node[box] (l43) at (3.6,6.0) {};

  \node[box] (l50) at (0,4.8) {};
  \node[box, draw=none] (l51) at (1.2,4.8) {};
  \node[box, draw=none] (l52) at (2.4,4.8) {};
  \node[box, draw=none] (l53) at (3.6,4.8) {};  

  \node[box] (l60) at (0,3.6) {};

  \node[box] (l70) at (0,2.4) {};



  \foreach \i in {1,2,3} {
    \draw[arrow3] (l1\i) -- (l2\i);
  }
  \foreach \i in {2,3} {
    \draw[arrow3] (l2\i) -- (l3\i);
  }
  \foreach \i in {0,3} {
    \draw[arrow3] (l3\i) -- (l4\i);
  }

  \foreach \i in {0,1,2,3} {
    \draw[arrow2] (l0\i) -- (l1\i);
  }

  \draw[arrow2] (l10) -- (l20);
  \draw[arrow2] (l12) -- (l20);
  \draw[arrow2] (l13) -- (l20);

  \draw[arrow2] (l20) -- (l30);
  \draw[arrow2] (l21) -- (l30);

  \draw[arrow2] (l30) -- (l42);
  \draw[arrow2] (l32) -- (l42);
  \draw[arrow2] (l33) -- (l42);

  \draw[arrow2] (l40) -- (l50);
  \draw[arrow2] (l42) -- (l50);
  \draw[arrow2] (l43) -- (l50);

  \draw[arrow2] (l50) -- (l60);
  
  \draw[arrow2] (l60) -- (l70);

\end{tikzpicture}
    \caption{Representation of the neural architecture of our GNN}
    \label{fig:neuralNetwork}
\end{figure}

\subsubsection{Training Algorithm} \label{sec:TrainingAlgorithm}
We train our GNN with backpropagation using the Adam optimizer \citep{kingma2014adam} and we set an initial learning rate of $0.001$.
The model is trained to minimize the binary cross-entropy loss
\begin{equation}
\mathcal{L} = - \frac{1}{N} \sum_{i=1}^{N} \Big[ y_i \log(\hat{y}_i) + (1 - y_i)\log(1 - \hat{y}_i) \Big]
\end{equation}
which measures the distance between predicted probabilities and true labels, penalizing wrong and non-confident predictions.
Training proceeds for a maximum of \num{2000} epochs.
If the validation loss does not improve for 40 consecutive epochs, we stop the training and retain the learnable parameters that yielded the minimal value of the loss function.

We generate the population of PPs for our training and validation sets by solving instances of our application problem using branch-and-price with full column generation.
For each solved PP instance, we collect the features and record them together with a label indicating the sign of the optimal value.
The resulting population is highly unbalanced due to two main factors.
The first is the skewness of the label distribution, caused by the fact that the PPs with a non-negative optimal value significantly outnumber those with a negative value.
The second is the over-representation of easy-to-classify PPs over hard-to-classify PPs.
The first ones have either a very positive or a very negative optimal value, while the second ones tend to have an optimal value close to zero.
While easy-to-classify PPs appear through the entire CG algorithm, hard-to-classify PPs tend to appear only in the later iterations of CG loops that converge slowly.
This is reflected in the population, in which we have a large share of easy-to-classify PPs mostly coming from the early CG iterations and a much smaller share of hard-to-classify PPs mostly coming from the late CG iterations.
To counter both unbalancing factors, we undersample PPs with a non-negative optimal value and PPs from the early CG iterations.

\section{Computational Study} \label{sec:computationalStudy}
In this section, we present our computational studies and the results that we obtained.
Here, we refer to our novel PCG strategy as the GNN strategy.
In Section \ref{sec:mainStudy}, we present our main computational study, where we evaluate the impact of our GNN strategy compared to those from the literature which were directly applicable to our application problem.
We implemented our GNN strategy as well as the ones from the literature and integrated them alternatively into the branch-and-price algorithm.
We used Python~3 to implement all our code and used the PyTorch and PyTorch Geometric libraries for the ML model.
We ran all experiments on a machine with an AMD Ryzen 9 5950X 16-Core Processor and two NVIDIA GeForce RTX 3090.
In Section \ref{sec:ablation} we present our ablation study.

\subsection{Instances}
We created three distinct sets of instances of our application problem using the instance generator from \citet{dall2023formation}.
The first set contains \num{20000} instances, which we used to create the population from which we sample the training set.
The second set contains \num{3200} instances, which we used to create the population from which we sample the validation set. 
We used the third set, which contains \num{1200} instances, to test our approach.
To obtain the populations of PPs from the instances of our application problems, we used the algorithm from \citet{hagn2024branch}.
We solved every instance with a time limit of 1 minute and created an observation from every PP solved.
We subsequently sampled the observations as described in Section \ref{sec:TrainingAlgorithm} to obtain balanced sets.

The instances have $7$ to $12$ PPs with an average of $9.2$.
The instances are characterized by the worker strength (WS), which is a measure of the number of workers available in comparison to the workload.
A formal definition of WS can be found in \citet{dall2023formation}.
Instances with a high WS tend to be easier to solve in terms of both finding a feasible solution and finding an optimal solution.
The lower the WS, the harder it becomes to solve the instances.
The WS varies from $0.4$ to $0.9$ with an incremental step of $0.1$

\subsection{Performance Evaluation} \label{sec:mainStudy}
We evaluate the performance of the different PCG strategies by analyzing their impact on the branch-and-price algorithm in finding optimal or good quality solutions.
We benchmark our approach against FCG, the PCG strategy from Gamache \citep{gamache1999column}, the one from Rothenb\"{a}cher \citep{rothenbacher2016branch}, and a random strategy.
For the Gamache and random strategies, we systematically varied the maximum number of negative columns and the probability of solving a PP, respectively.
In Section \ref{sec:tightRuntime}, we present the results of our computational study in which we compare the performance of the different PCG strategies under a tight time limit. 
In Section \ref{sec:extendedRuntime}, we present our extended-runtime study on a subset of test instances.

Note that we refrain from applying our GNN strategy in the following two cases.
First, in the root node of the branching tree because we observed that it is not beneficial.
Second, when the PP forbids the generation of certain columns since our input graph does not model this type of constraint.
In these two cases we use full column generation.

We calculate the overhead of a PCG strategy as the percentage of time spent on additional computations required by its application, relative to the total computational time of the algorithm.
The overhead of the calculations required to apply the traditional PCG strategy is negligible.
Conversely, the overhead induced by our strategy is $5.43$\%, which is significant.
Specifically, the average overhead to collect the features and build the partially bipartite input graph is $2.36$\% while the average overhead to make the actual predictions is $3.07$\%.

\subsubsection{Comparison with Traditional Methods}\label{sec:tightRuntime}
In our main study, we solve all test instances under a time limit of 1 minute.
We let the branch-and-price algorithm run for at most 45 seconds.
If it cannot find any optimal solution, we use the generated columns to run the early termination heuristic from Section \ref{sec:earlyTerminationHeuristic} for at most 15 seconds.
In case the early termination heuristic does not return any feasible solution, we say that the algorithm failed to solve the instance.

Table \ref{tab:comparisonLiterature} presents a comparison between our GNN strategy, indicated as ``GNN'', and the PCG strategies from the literature.
\begin{table}[h]
    \centering
    \small
    \begin{tabular}{lcccccccccccccccccccccccccccc}
    \toprule
     & Solved & Optimal & Gap\textsubscript{H} & Gap\textsubscript{B} & RMSD(gap) \\
    \midrule
FCG & 97.9\% & 61.8\% & 8.51\% & 7.63\% & 0.064 \\
Rothenbaecher & 98.0\% & 61.5\% & 7.80\% & 7.30\% & 0.060 \\
Gamache 1 & 98.1\% & 61.4\% & 8.22\% & 7.55\% & 0.062 \\
Gamache 2 & 98.0\% & 61.4\% & 8.23\% & 7.70\% & 0.063 \\
Gamache 3 & 98.1\% & 61.4\% & 8.58\% & 7.74\% & 0.064 \\
Gamache 4 & 98.0\% & 61.6\% & 8.58\% & 7.84\% & 0.066 \\
Gamache 5 & 98.0\% & 61.6\% & 8.44\% & 7.68\% & 0.064 \\
Gamache 6 & 98.0\% & 61.4\% & 8.55\% & 7.85\% & 0.066 \\
Gamache 7 & 98.0\% & 61.6\% & 8.50\% & 7.82\% & 0.066 \\
Gamache 8 & 97.9\% & 61.6\% & 8.65\% & 7.94\% & 0.068 \\
Gamache 9 & 98.0\% & 61.6\% & 8.49\% & 7.77\% & 0.066 \\
Gamache 10 & 98.0\% & 61.6\% & 8.49\% & 7.78\% & 0.066 \\
Random 0.1 & 97.9\% & 61.7\% & 8.22\% & 7.38\% & 0.061 \\
Random 0.2 & 97.7\% & 61.7\% & 8.42\% & 7.68\% & 0.065 \\
Random 0.3 & 98.1\% & 61.6\% & 8.14\% & 7.32\% & 0.060 \\
Random 0.4 & 98.3\% & 61.9\% & 7.64\% & 7.00\% & 0.057 \\
Random 0.5 & 97.9\% & 61.7\% & 8.34\% & 7.54\% & 0.064 \\
Random 0.6 & 97.9\% & 61.7\% & 8.17\% & 7.44\% & 0.062 \\
Random 0.7 & 98.1\% & 61.4\% & 8.19\% & 7.52\% & 0.062 \\
Random 0.8 & 97.8\% & 61.8\% & 8.56\% & 7.74\% & 0.066 \\
Random 0.9 & 97.8\% & 61.7\% & 8.35\% & 7.57\% & 0.063 \\ 
GNN & \textbf{98.5}\% & \textbf{62.0}\% & \textbf{7.26}\% & \textbf{6.36}\% & \textbf{0.049} \\
    \bottomrule
    \end{tabular}
    \caption{Performance comparison of the branch-and-price algorithm applying the different PCG strategies}
    \label{tab:comparisonLiterature}
\end{table}
Column ``Solved'' reports the share of instances for which the algorithm could find a feasible solution, while column ``Optimal'' shows the share of instances that were solved to optimality. 
The columns ``Gap\textsubscript{H}'' and ``Gap\textsubscript{B}'' show the average heuristic optimality gap and the average optimality gap against the best known solution.
For these columns, we only consider those instances for which at least one of the different approaches could not find an optimal solution.
Given an instance, we calculate the heuristic optimality gap as $\frac{UB-LB}{UB}$ and the optimality gap against the best known solution as $\frac{BUB-LB}{BUB}$ where $UB$ and $LB$ are the value of the solution and the best lower bound found by the algorithm, respectively, and $BUB$ is the value of the best solution.
When one of the algorithms cannot find any solutions for a specific instance, we consider Gap\textsubscript{H} and Gap\textsubscript{B} to be $100\%$.
We calculate the Rooted Mean Squared Deviation (RMSD) between $BUB$ and $LB$ as 
\begin{equation}
    \sqrt{\frac{\sum_{i=1}^N(UB-LB)^2}{N}}
\end{equation}
where $N$ is the number of instances.
We report this result in in column ``RMSD(gap)''.
We can easily see that our approach clearly outperforms those from the literature.
It can solve more instances, it can find more optimal solutions and the average quality of the suboptimal solution is higher.
Furthermore, it tends to be more stable and produces solutions whose value deviate the least from the optimum in comparison with the other methods, as can be deduced from the low RMSD.

In Table \ref{tab:comparisonHard}, we highlight the performance on the instances with a WS of $0.4$ and $0.5$, which tend to be the hardest to solve.
We report the performance of the FCG and of our approach in the corresponding columns.
In column ``$\Delta$'', we show the difference for the relative metric.
As expected, both algorithms perform worse on the harder instances.
However, we can see that our approach has an even more positive impact on the resolution of hard instances as one can see comparing the two ``$\Delta$'' columns.
In fact, the share of solved and optimally solved instances increases even more for the hard instances.
Similarly,  the optimality gaps and the RMSD have an even greater decrease.

\begin{table}[h]
    \centering
    \small
\begin{tabular}{lccccccc}
\hline
 & \multicolumn{3}{c}{All $WS$} &  & \multicolumn{3}{c}{ $WS \in \{0.4, 0.5\}$}  \\ \cline{2-4} \cline{6-8} 
 & FCG & GNN & $\Delta$ &  & FGC & GNN & $\Delta$ \\ \hline
Solved & 97.9\% & 98.5\% & +0.64\% &  & 93.6\% & 95.6\% & +2.00\% \\
Optimal & 61.8\% & 62.0\% & +0.36\% &  & 20.7\% & 21.2\% & +2.38\% \\
Gap\textsubscript{H} & 8.60\% & 7.36\% & $-$15.0\% &  & 9.19\% & 6.83\% & $-$25.6\% \\
Gap\textsubscript{B} & 7.73\% & 6.47\% & $-$17.0\% &  & 8.57\% & 6.07\% & $-$29.3\% \\
RSMD & 0.063 & 0.049 & $-$22.6\% &  & 0.092 & 0.068 & $-$26.5\% \\ \hline
\end{tabular}
    \caption{Comparison showing the stronger performance gains of our GNN strategy relative to FCG on hard instances.}
    \label{tab:comparisonHard}
\end{table}

\subsubsection{Extended-Runtime Study}\label{sec:extendedRuntime}
In this study, we extended the time limit to 5 and 10 minutes to observe the long-run performance of our GNN strategy.
In both cases, we reserve the last 2 minutes for the early termination heuristic.
Due to time limitations, we restricted the test set for this study to 120 instances.
We selected the restricted test set to maintain a similar distribution of instance characteristics as the original set.
In this way, the results remain representative of the full spectrum of the problem.
To run this study, we retrained our ML model with samples that we collected by letting the branch-and-price algorithm run for a longer time.
For the 5-minutes study, in which the branch-and-price runs for at most 3 minutes, we collect samples under a time limit of 3 minutes.
For the 10-minutes study, in which the branch-and-price runs for at most 8 minutes, we collect samples under a time limit of 5 minutes.
This puts our predictor under unfavorable conditions since it has to make predictions for PPs stemming from unseen circumstances.
The total number of training samples is the same as in the experiments presented in Section \ref{sec:tightRuntime} for both studies.

\begin{table}[h]
    \centering
    \small
\begin{tabular}{llllll}
\toprule
 & Solved & Optimal & Gap\textsubscript{H} & Gap\textsubscript{B} & RMSE(Gap) \\
\midrule
FCG & \textbf{100.0}\% & \textbf{70.1}\% & 8.99\% & 6.30\% & \textbf{0.013} \\
Rothenbaecher & \textbf{100.0}\% & \textbf{70.1}\% & \textbf{7.62}\% & 6.31\% & 0.014 \\
Gamache& \textbf{100.0}\% & 70.0\% & 7.71\% & 6.28\% & \textbf{0.013} \\
Random & 99.9\% & 70.0\% & 8.26\% & 6.51\% & 0.016 \\
GNN & \textbf{100.0}\% & \textbf{70.1}\% & 7.78\% & \textbf{6.22}\% & 0.014 \\
\bottomrule
\end{tabular}
\caption{Results of the 5 minutes extended-runtime study}
\label{tab:longRun5}
\end{table}

\begin{table}[h]
    \centering
    \small
\begin{tabular}{llllll}
\toprule
 & Solved & Optimal & Gap\textsubscript{H} & Gap\textsubscript{B} & RMSE(Gap) \\
\midrule
FCG & \textbf{100.0}\% & \textbf{70.9}\% & 8.38\% & 5.86\% & \textbf{0.012} \\
Rothenbaecher & \textbf{100.0}\% & \textbf{70.9}\% & \textbf{7.34}\% & \textbf{5.82}\% & \textbf{0.012} \\
Gamache & \textbf{100.0}\% & \textbf{70.9}\% & \textbf{7.34}\% & 5.84\% & \textbf{0.012} \\
Random & 99.8\% & 70.8\% & 8.00\% & 6.41\% & 0.018 \\
GNN & \textbf{100.0}\% & \textbf{70.9}\% & 7.87\% & 5.86\% & \textbf{0.012} \\
\bottomrule
\end{tabular}
\caption{Results of the 10 minutes extended-runtime study}
\label{tab:longRun10}
\end{table}

Table \ref{tab:longRun5} and Table \ref{tab:longRun10} show the results of the 5 and 10 minute extended-runtime study, respectively.
In both tables, we average the results of the different settings of the Gamache and Random PCG strategies.
In the 5-minutes study, our approach can find the best quality solutions on average compared to the other PCG strategies, as one can observe in column ``Gap\textsubscript{B}''.
However, the Rothenb{\"a}cher strategy can find the best upper bounds on average, as can be deduced from column ``Gap\textsubscript{H}''. 
In the 10-minutes study, the performance of our approach is outperformed by the Gamache and Rothenb{\"a}cher strategies.
Despite not being the most performative PCG strategy in all settings, our approach constantly outperforms FCG.

We identified the following reasons for the poorer performance of our GNN strategy in the extended-runtime study compared to the study in Section \ref{sec:tightRuntime}.
First, letting the branch-and-price run for longer time results in a more heterogeneous population of PPs, which makes the predictions more challenging.
Second, the majority of columns produced by the CG after the late stages of the branch-and-price algorithm tend to have a reduced cost close to 0, which are harder to classify for our predictor.
If we let the branch-and-price run for a longer time, we will encounter more of these columns, which renders our GNN strategy less effective.
Lastly, the training set lacks samples from runs longer than 5 minutes.
This affects the ability of our ML model to make correct predictions in the 10-minutes study.

\subsection{Ablation Study} \label{sec:ablation}
To validate the architecture of our ML model, we performed an ablation study by altering or removing specific components.
Our ablation study consists of 4 alternative architectures.
First, we alter the convolutional layers, where we replace the graph isomorphism networks of our original architecture with graph attention networks \citep{brody2021attentive}.
Second, we remove the positional encoding in the supplementary nodes to evaluate its relevance.
Finally, we simplify the structure of our input graph making it an arbitrary directed graph instead of a partially bipartite graph.
We obtain this by removing the supplementary nodes and edges, and including all the information in the transportation graph.
We detail the two simplified input graphs of our ablation study in the following paragraph.

\paragraph{Simplified Input Graphs}
Since the simplified input graphs in our ablation study only consist of the transportation graph, the latter requires additional features to incorporate the information originally contained in the supplementary nodes or edges.
We recall that the supplementary nodes and edges incorporate the time components $\zeta_\tau \in \mathcal{T}$ and the time-window parameters $ES_i$, $LF_i$, $LF^\mathrm{e}_i$, $EF_i$, $EFQ_{i,q}$, $LSQ_{i,q}$ for each task $i \in \mathcal{I}_q$.
For a node corresponding to task $i$, we define the vector of relevant time components $\mathcal{Z}_i = [ \zeta_\tau \mid ES_i \leq \tau \leq LF^\mathrm{e}_i ]$.
Similarly, we define the set of relevant time components of an arc corresponding to the pair of tasks $(i,j)$ in $\mathcal{I}_q$ as $\mathcal{Z}_{i,j} = \left[ \tau \in \mathcal{T} \mid EFQ_{i,q} \leq t \leq LSQ_{i,q} \right]$.
In the first simplified input graph, we append statistics on the vector of relevant time components $\mathcal{Z}_i$ and $\mathcal{Z}_{i,j}$ to the feature vector of each task $i \in \hat{\mathcal{V}}^T_q$ and arc $(i,j) \in \hat{\mathcal{A}}^T_q$, respectively.
The statistics we consider are minimum, maximum, summation, mean, standard deviation, kurtosis, and skewness of the vector of relevant time components.
In the second simplified input graph, we directly append the corresponding vector of relevant time components to the features of each node and arc.
We use a padding strategy to ensure that the vectors of relevant time components have a fixed size.
In both simplified models, the time-window parameters become features of the transportation nodes.
This is straightforward since the time-window parameters are defined on the set of tasks.
Unlike in our partially bipartite input graph, the values of time-window parameters are included as features in an explicit fashion.
For this reason, we apply a custom normalization to the time window parameters, subtracting the first meaningful point in time of the specific PP and dividing by the maximum number of points in time.
Since the simplified graphs do not contain the supplementary nodes and edges, we replace the bipartite convolutional layers with in-convolutional layers.

\begin{figure}
    \centering
    \input{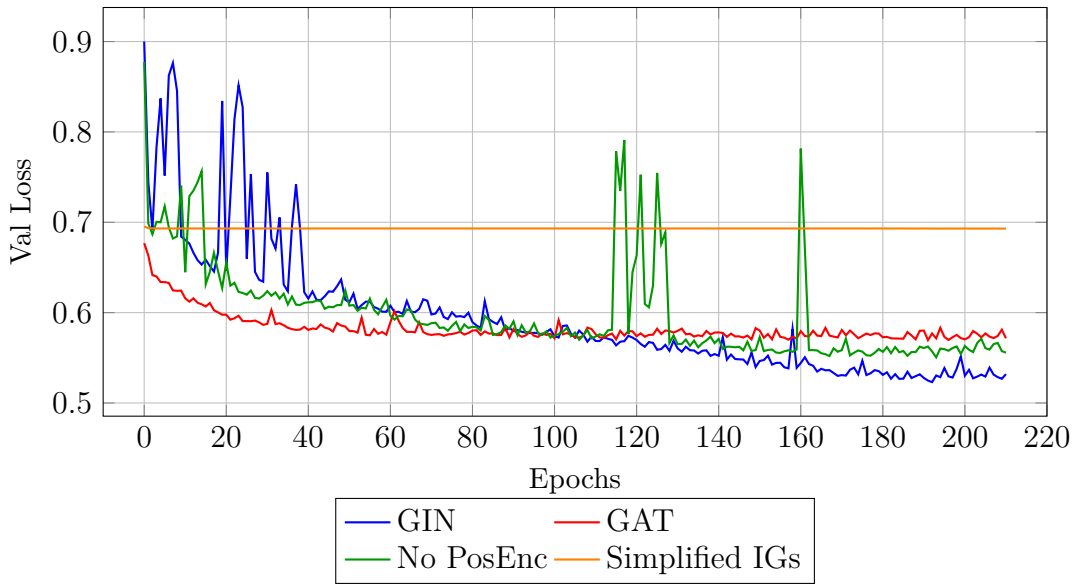}
    \caption{Evolution of the validation loss in our ablation study}
    \label{fig:valLoss}
\end{figure}

We report the evolution of the validation loss of our original model and the alternative models across the training epochs in Figure \ref{fig:valLoss}.
Each curve corresponds to a different model, with the global minimum indicating its best performance.
Our original model, labeled ``GIN'', outperforms all other models since its global minimum is the lowest overall.
The curves corresponding to the simplified input graphs (simplified IGs) are very similar and they are indistinguishable on the graph.
For both simplified input graphs, the validation loss plateaus prematurely at a high value, indicating poor performance.
This validates the more complex but more effective structure of our partially bipartite input graph.
In particular, the structure of our partially bipartite input graph has two main advantages: (i) it does not require to arbitrarily aggregate the value of the dual values to fit them directly into the transportation graph and (ii) it makes it possible to implicitly model the time window parameters considering their semantics rather than their value.
The validation loss of the model using attention mechanisms, labeled ``GAT'', is more stable and converges faster, but stabilizes at a higher value than our model.
The model without positional encoding (No PosEnc) also converges at a higher values compared to our model.
This shows that the positional encoding in the supplementary nodes significantly improves overall performance.

\section{Conclusions}
In this paper, we have studied an ML-based acceleration for the branch-and-price algorithm to solve the team formation and routing problem.
Specifically, we developed a novel partial column generation (PCG) strategy that uses a prediction model.
Our approach uses graph neural networks (GNNs) to dynamically select which pricing problems (PPs) to solve in each column generation (CG) iteration based on the likelihood of producing columns with a negative reduced cost.
We proposed a partially bipartite graph to represent the pricing problem instances, which constitute elementary shortest path problems with resource constraints.

Our computational study shows that our method outperforms the original full CG algorithm and traditional PCG strategies from the literature under a tight time limit.
Specifically, the branch-and-price algorithm is able to find more optimal solutions, lower optimality gaps for suboptimal solutions, and more feasible solutions overall when applying our GNN-based PCG strategy.
Performance improvement is even more evident for hard instances characterized by low resource availability.
The ablation study validates our architectural choices for the deep neural model.
In particular, it highlights better performance for the partially bipartite input graph over an arbitrary input graph.
One main limitation of our approach is long-term performance degradation under extended time limits.
This is due to the increasing hardness of the PP predictions and the growing heterogeneity of the PPs associated with the later stages of the branch-and-price.
To mitigate this effect, our research could be extended by devising a metric that determines when to switch from our ML-based approach to other PCG strategies.
Furthermore, refinements of the deep learning model to improve performance in the late stages could be explored.

As future research, we suggest the application of our proposed GNN-based PCG strategy to other problems.
We identified rich vehicle routing problems with complex PPs as a potential area to apply our method.

\bibliographystyle{elsarticle-harv} 
\bibliography{bibliography}

\appendix
\section{Feasibility Check} \label{app:feasibilityCheck}
We define the set of team routes that make up an RMP integer solution $\bar{\lambda}$ as $\bar{\mathcal{R}} = \left\{ (r,q) \in \mathcal{R} : \bar{\lambda}^r_q = 1\right\}$.
We define a graph containing a node for each team route $(r,q) \in \bar{\mathcal{R}}$ plus a source node $o$ and a sink node $o'$.
For each team route $(r,q) \in \bar{\mathcal{R}}$, the set of predecessors is $\bar{\mathcal{R}}^r_{-} = \left\{(\tilde{r}, \tilde{q}) \in\bar{\mathcal{R}}: \mathrm{tr}^{\tilde{r},\tilde{q}} \leq \mathrm{tl}^{r,q}\right\} \cup \left\{o\right\}$ and the set of successors is $\bar{\mathcal{R}}^r_{+} =\left\{(\tilde{r}, \tilde{q}) \in\bar{\mathcal{R}}: \ \mathrm{tl}^{\tilde{r},\tilde{q}} \geq \mathrm{tr}^{r,q}\right\} \cup \left\{o'\right\}$.
We define integer flow variables $x_k^{r,\varrho}$ that indicate the number of workers with skill level $k$ consecutively assigned to $r$ and then to $\varrho$ or leaving or returning to the depot if $r=o$ or $\varrho=o'$, respectively.
To simplify the notation and render the model independent of the index $q$ referring to the working profile, we define a parameter $\beta^r_k = \xi_{q,k}$ for each team route $(r,q) \in \bar{\mathcal{R}}$
The formulation of the feasibility problem is as follows.
\begin{samepage}
\begin{align}
& & & \sum_{\rho \in \bar{R}^{r}_{-}} \sum_{\kappa = k}^{K} x^{\rho,r}_{\kappa} \geq \beta^r_k & \forall r \in \bar{R}, \forall k \in \mathcal{K} \label{assignEnoughWorkersFeas}\\
& & & \sum_{\rho \in \bar{R}^r_{-}} x^{\rho,r}_k = \sum_{\varrho \in \bar{R}^r_{+}} x^{r,\varrho}_k & \forall r \in \bar{R}, \forall k \in \mathcal{K} \label{flowConservationFeas}\\
& & & \sum_{r \in \bar{R}} x^{o,r}_k = n_k & \forall k \in \mathcal{K} \label{srcFeas}\\
& & & \sum_{r \in \bar{R}} x^{r,o'}_k = n_k & \forall k \in \mathcal{K} \label{snkFeas}\\
& & & x^{r,\varrho}_k \in \mathbb{Z}_{\geq 0} & \forall r \in \bar{R} \cup \left\lbrace o,o' \right\rbrace, \forall \varrho \in \bar{R}^r_{+}, \forall k \in \mathcal{K} \label{dom:Xfeas}
\end{align}
\end{samepage}
where $n_k$ is the number of workers available with exact skill level $k$.






\end{document}